\documentclass[12pt]{article}

\usepackage{times}
\usepackage{url}
\usepackage{amsmath}
\usepackage{amsfonts}
\usepackage{graphicx}
\usepackage{comment}
\usepackage{booktabs}
\usepackage{threeparttable}
\usepackage{multirow}
\usepackage{mdwtab}
\usepackage{textcomp} 
\usepackage{xurl}
\usepackage[normalem]{ulem}
\newcommand\redout{\bgroup\markoverwith
{{\rule[0.5ex]{2pt}{1pt}}}\ULon}

\usepackage{xcolor}

\usepackage{caption}
\newenvironment{sciabstract}{%
\begin{quote} \bf}
{\end{quote}}

\title{Transferable Human Mobility Network Reconstruction with neuroGravity}

\author
{Jinming Yang$^{1,2}$, Shaoyu Huang$^{1}$, Zongyuan Huang$^{1}$, Yaohui Jin$^{1,2\ast}$,\\ Xiaokang Yang$^{1\ast}$, Marta C. Gonz\'{a}lez$^{3,4,5}$, Yanyan Xu$^{1,2\ast}$\\
\\
\normalsize{$^{1}$MoE Key Laboratory of Artificial Intelligence, AI Institute,}\\
\normalsize{Shanghai Jiao Tong University, Shanghai 200240, China}\\
\normalsize{$^{2}$Data-Driven Management Decision Making Lab,}\\
\normalsize{Shanghai Jiao Tong University, Shanghai 200240, China}\\
\normalsize{$^{3}$Department of City and Regional Planning, University of California,}\\
\normalsize{Berkeley, CA 94720, USA}\\
\normalsize{$^{4}$Energy Technologies Area, Lawrence Berkeley National Laboratory,}\\
\normalsize{Berkeley, CA 94720, USA}\\
\normalsize{$^{5}$Department of Civil and Environmental Engineering, University of California,}\\
\normalsize{Berkeley, CA 94720, USA}\\
\\
\normalsize{$^\ast$To whom correspondence should be addressed. E-mail: yanyanxu@sjtu.edu.cn.}
}

\date{}

\begin{document} 
\baselineskip24pt
\maketitle 

\begin{sciabstract}
Accurate modeling of human mobility is critical for tackling urban planning and public health challenges. In undeveloped regions, the absence of comprehensive travel surveys necessitates reconstructing mobility networks from publicly available data. Here we develop neuroGravity, a physics-informed deep learning model that reliably reconstructs mobility flows from limited observations and transfers to unobserved cities. Using only urban facility and population distributions, we find that neuroGravity's regional representations strongly correlate with socioeconomic and livability status, offering scalable proxies for costly surveys. Furthermore, we uncover that spatial income segregation plays a key role in model transferability: mobility networks are most reliably reconstructed when target cities share similar segregation levels with the source. We design an index to quantify this segregation and accurately predict transferability. Finally, we generate mobility flow proxies for over 1,200 cities worldwide, highlighting neuroGravity's potential to mitigate critical data shortages in resource-limited, underdeveloped areas.

\end{sciabstract}

\section*{Introduction}

Rapid urbanization in contemporary society is restructuring urban infrastructures, economies, and societies, complicating the interaction between human and built environments~\cite{batty2008size,bettencourt2013origins,verbavatz2020growth}. Understanding human mobility patterns is critical for addressing challenges in urban planning, traffic management, and epidemic disease control~\cite{gonzalez2008understanding, xu2018planning,vazifeh2018addressing,buckee2020aggregated,tian2020investigation,barbosa2018human}. The traditional approach collects residents' travel demand via manual surveys~\cite{CTPP2024}; however, this method is associated with substantial economic and time costs. With the popularization of information and communication technology in the past 20 years, people's mobility trajectories can be collected and aggregated to mobility flows~\cite{toole2015path}. Whereas in undeveloped regions, especially the Global South, costly travel surveys and limited infrastructures result in the scarcity of reliable mobility information~\cite{DigitalEconomy2024}.

As a fundamental driving force of urban dynamics, human mobility is highly associated with multiple aspects, from urban structure to socioeconomic status~\cite{moro2021mobility,xu2023urban,nilforoshan2023human}. As early as 1946, Zipf proposed the classical gravity model, which utilized population data and travel distance to estimate inter-city travel flows~\cite{zipf1946p}. Simini et al. proposed a radiation model, a parameter-free alternative to the gravity law, to estimate mobility fluxes~\cite{simini2012universal}. Due to its simplicity and practicality, the gravity model remains a primary tool to estimate human mobility networks to this day~\cite{ganin2017resilience,kraemer2017spread}. In the last decade, advanced machine learning techniques have been adopted to produce more realistic flows and to develop more sophisticated physical mobility models by integrating a broader range of influential features~\cite{simini2021deep,song2024enhancing,zhao2024predicting,cabanas2025human}. 
For example, Deep Gravity~\cite{simini2021deep} uses a neural network to estimate movement probabilities, but fundamentally requires known total outflows to generate absolute volumes. This limitation renders it incapable of reconstructing mobility networks in cities lacking prior mobility data (Supplementary Fig.~8).

Fig.~1a schematically depicts the applicability of physical and deep learning methods for the reconstruction of mobility networks. Physical models, typified by the gravity law, capture the holistic distribution of mobility flow. They are completely interpretable, but with relatively weak reconstruction performance as a result of limited input information and parameters. When abundant regional characteristics are available, deep learning methods excel in capturing complex flow patterns, albeit at the expense of interpretability. Moreover, transferability is another important aspect when reconstructing mobility networks for cities without observations. In general, physical models demonstrate greater transferability than deep learning models primarily due to the latter's susceptibility to overfitting. This advantage is especially pronounced when transferring models between cities with distinct characteristics.

Building on the respective strengths of physical and deep learning methods, we develop \textbf{neuroGravity}, a hybrid model that estimates the absolute values of mobility flow between two given regions. NeuroGravity integrates the classical gravity model and graph neural networks (GNNs) within a unified framework, enabling it to perform robustly even in data-scarce scenarios. The mobility network in Boston, for instance, containing 51,786 edges (representing origin-destination pairs) among 250 nodes (representing ZIP code tabulation area), can be successfully reconstructed ($R^2=0.77$) even when 1\% (496 of 51,786) of the internal origin-destination pairs are observed.

By assimilating mobility interactions, neuroGravity’s regional embeddings implicitly capture socioeconomic dynamics absent from OpenStreetMap. Using a gradient boosting machine (GBM), these embeddings effectively predict key socioeconomic and livability indicators, including income, education, carbon footprint, nitrogen dioxide, and radius of gyration (Fig.~1b). Despite varying across urban structures, they offer cost-effective proxies for traditional surveys and environmental monitoring.

Given the economic barriers to conducting comprehensive travel surveys in the Global South, generating mobility networks from publicly accessible data is crucial. We show that neuroGravity addresses this zero-shot flow generation challenge effectively, leveraging models trained in data-rich cities like Boston to construct flow networks for unobserved cities (Fig.~1c). Crucially, we find that spatial income segregation governs this transferability. Target cities sharing similar segregation profiles with the source city yield higher accuracy. Inspired by Chodrow~\cite{chodrow2017structure}, we introduce a spatial segregation index to quantify this effect, revealing that neuroGravity's zero-shot transferability can be reliably predicted with a simple linear model.

Finally, we applied the neuroGravity model to generate mobility flows for over 1,200 cities and regions on a global scale (Fig.~1d). We also illustrate that generated flows can be adopted to drive susceptible-exposed-infectious-recovered (SEIR) epidemiological models~\cite{chang2021mobility, lai2020effect}, demonstrating accurate pandemic dynamics comparable to those driven by observed flows. These results underscore the potential of neuroGravity in facilitating the application of mobility models in diverse urban science and public health scenarios, even in the absence of extensive local data.

\section*{Results}
\subsection*{Mobility flow reconstruction with minimal observation}

To tackle the challenge of reconstructing human mobility networks from limited data, we developed neuroGravity, a physics-informed deep learning framework that integrates the gravity law with graph representation learning. The pipeline begins by extracting high-dimensional demographic and built-environment features for each geographic region (Fig. 2a). Given the inherent sparsity of urban mobility, a connection predictor evaluates these features to identify plausible origin-destination (OD) pairs, establishing the base network topology (Fig. 2b). Flow estimation then proceeds through two coupled stages: a deep-parameterized ``meta-Gravity'' component generates a physically grounded base flow estimation (Fig. 2c), which is subsequently refined by an edge-enhanced graph neural network (Fig. 2d). By operating in a logarithmic space, neuroGravity inherently preserves the multiplicative nature of the gravity law while capturing complex, context-specific spatial dependencies (a detailed formulation is provided in the Methods section).

According to the way of collecting the flow observations, the task of reconstructing human mobility networks can fall into the three scenarios, (i) random observations, (ii) observations from or to a part of nodes, and (iii) internal observations, which means only the internal flow (movements that both originate and terminate within the observed subset of regions) of a small subset of regions can be observed, as presented in Supplementary Figs.~1 and 7. Here we focus on the third scenario, which is the most practical but the most challenging. We evaluate neuroGravity’s flow reconstruction across six cities: Boston, Los Angeles (LA), San Francisco Bay Area (SF Bay), Porto, Bogot\'a, and Riyadh. Using internal flows from a fraction (e.g., 10\%) of randomly selected regions as observations (denoted as 10\% observation ratio; Supplementary Fig.~7), the model estimates flows for all remaining origin-destination (OD) pairs. Additionally, to resolve limitations in the original Deep Gravity model, we introduce DG++, an enhanced regression variant with modified output and loss functions (Supplementary Section 4.3).

Figs.~3a-c highlight the reconstruction performance in the three U.S. cities under a 10\% observation ratio, which covers 0.7-1.2\% of all flow links. To enhance clarity, we display only the top 30\% of flow links, using color and width to represent flow percentile rankings relative to the ground truth, from the highest (100\%) down to the 70th percentile. Fig.~3a presents the results of Boston, starting from the observed mobility flows, to the flows generated by the gravity model, neuroGravity, and the actual flow from CDRs. Figs.~3b-c further illustrate the models' performances for LA and SF Bay. Our results reveal that the flow distributions reconstructed by neuroGravity closely mirror the actual mobility patterns, effectively capturing major flow arteries and high-density areas. 
In contrast, the gravity model tends to centralize flows around the city center, resulting in an oversimplified representation that lacks the nuanced detail of peripheral connections.

Supplementary Fig.~9 further demonstrates the performance of additional models in the six cities. Although GNN shows improved coverage of peripheral flows compared to the gravity model, it struggles to differentiate between high- and low-volume flow links. DG++ demonstrates reasonable accuracy in reconstructing mobility flows for cities with homogeneous spatial interactions, such as the San Francisco Bay Area. However, it becomes susceptible to overfitting and fails to generalize in cities with more diverse flow patterns.

Figs.~3d-e are scatter plots of the estimated flows (under the 10\% observation ratio) versus actual flows for Boston using the gravity model and neuroGravity, respectively. The gravity model achieves an \( R^2 \) of 0.59 and a common part of commuters (CPC)~\cite{lenormand2016systematic} of 0.63 (Fig.~3d), suggesting moderate alignment with actual flows but substantial variances, particularly at higher flow values. In contrast, neuroGravity achieves a higher \( R^2 \) of 0.77 and a CPC of 0.73 (Fig.~3e). This is accompanied by more compact alignment with the \( y=x \) line, narrower boxplot widths, and tighter 75\% confidence intervals. 
Additional results comparing more baselines across six cities in Supplementary Fig.~10 also highlight neuroGravity's superior accuracy and stability.

Figs.~3f-g present the stability and consistency of neuroGravity's flow reconstruction across different observation ratios, as evaluated by the median $R^2$ and CPC metrics over 30 independent runs, each using randomly sampled observations. As the observation ratio increases, neuroGravity consistently outperforms the gravity model in both \( R^2 \) and CPC, highlighting its robustness in reconstructing mobility flows. It's noteworthy that the gravity model's performance declines with the increase in observation ratio. The main reason is that the gravity model captures broad flow distributions well but struggles with precise flow estimates for specific links. As the observation ratio increases, the remaining test samples may deviate from the average, leading to lower \( R^2 \) values for the gravity model.

More results are presented in Supplementary Fig.~11. Among these, DG++ outperforms both the gravity model and GNN in most cases. Notably, under relatively high observation ratios, DG++ approaches neuroGravity in LA, demonstrating great potential when ample data is available. However, its performance is markedly unstable across various geographical contexts, suffering failures in other cities (e.g., Porto and Bogot\'a). Consequently, the GNN emerges as the strongest baseline in terms of average \( R^2 \) across all cities under the most challenging 10\% observation scenario (Supplementary Table~4-6). NeuroGravity substantially outperforms this best-average baseline, achieving a 38\% \( R^2 \) improvement in average, which underscores the advantage of neuroGravity in adaptability and stable performance in diverse urban environments.

\subsection*{Projecting regional socioeconomic status from neuroGravity embeddings}

Beyond flow reconstruction, neuroGravity generates regional embeddings that capture critical urban functional structures. We demonstrate that these representations help estimate socioeconomic and livability indicators, including household income, educational attainment, carbon footprint, nitrogen dioxide concentration, and radius of gyration ($Rg$), offering scalable proxies to complement resource-intensive surveys. 

In Fig.~4a, we present the 2-dimensional UMAP projection~\cite{becht2019dimensionality} of neuroGravity's node embedding space for Boston neighborhoods, where each point corresponds to a ZCTA and is colored with median household income. This visualization reveals distinct clustering patterns that align with household income levels, despite the fact that income data were not incorporated into neuroGravity. Notably, ZCTAs with comparable income levels, such as the pairs zone1-zone3 and zone2-zone5, form cohesive clusters in the embedding space, regardless of their geographical distances. Conversely, geographically adjacent zones of different income levels (e.g., zone2, zone3, and zone4) are distinctly separated in the embedding space. 

Fig.~4b examines pairwise income gaps as a function of their embedding distances. Each box represents the distribution of income gaps between pairs of regions that share a similar embedding distance. A pronounced positive correlation is observed between the embedding distance and the median income disparity (denoted by the red line), indicating that spatial proximity within the embedding space is associated with socioeconomic homogeneity. This pattern reinforces the observation that neuroGravity's embeddings implicitly capture socioeconomic distinctions among regions.

We then quantify neuroGravity embeddings' capacity to estimate regional indicators by training a GBM model with a combined set of OSM attributes and node embeddings.
Results for Boston (Fig.~4c) show high \( R^2 \) values for traffic-related indicators, such as household carbon footprint (\( R^2 = 0.73 \pm 0.02 \)), NO\textsubscript{2} concentration (\( R^2 = 0.78\pm 0.01 \)), and \( R_g \) (\( R^2 = 0.76 \pm 0.02 \)). It also achieves modest performance for household income (\( R^2 = 0.42 \pm 0.02 \)) and college degree ratio (\( R^2 = 0.35\pm 0.03 \)). Fig.~4d confirms that combining OSM attributes with neuroGravity embeddings outperforms separate training, demonstrating that the embeddings encapsulate additional mobility-related information that enhances socioeconomic inference.

It is noteworthy that the impact of incorporating embeddings into the input feature set varies across cities. The improvement is most pronounced in Boston, moderate in SF Bay, and least noticeable in LA. Moreover, both OSM features and neuroGravity embeddings show weak performance in estimating the radius of gyration $Rg$ in LA ($R^2=-0.07 \pm 0.05$) and SF Bay ($R^2=0.13 \pm 0.05$). These variations can be attributed to differences in urban structure~\cite{xu2023urban}. In monocentric cities like Boston, residents from suburban areas often travel longer distances to centralized urban hubs for work or services, resulting in centripetal mobility patterns that correlate strongly with socioeconomic metrics. In contrast, the dispersed urban layout in polycentric cities (LA and SF Bay) leads to more heterogeneous travel patterns. The functionalities of different regions are more mixed, and socioeconomic status exhibits more complex relationships with residents' mobility behavior.

To quantify individual feature contributions, we applied PCA~\cite{abdi2010principal} to condense neuroGravity embeddings into four components (\texttt{neuroGravity\_f0-3}), and then conducted SHapley Additive exPlanations (SHAP) analysis~\cite{scott2017unified} across three cities and five indicators (15 tests in total) to assess variable contributions. Fig.~4e presents the 25 most important variables sorted by their median relative contributions across the 15 tests. Two neuroGravity features are ranked first and third, confirming their substantial contributions to socioeconomic inference. Additional evaluations are provided in Supplementary Section 6.2 (Supplementary Figs.~12-16).

\subsection*{Cross-city mobility flow generation and the role of spatial income segregation}

Generating mobility flow in cities without any observations holds tremendous potential for urban planning, pandemic modeling, and infrastructure optimization. Here, we evaluate the cross-city flow generation capability of neuroGravity using mobility flow derived from CDR, location-based services (LBS), and census commuting flow data across various cities and regions worldwide.  To ensure robust flow generation, we employ a model-assembling technique, combining multiple base models trained from independently sub-sampled mobility networks (Supplementary Fig.~17). 

Fig.~5a presents mobility flows generated by a neuroGravity model trained using Boston CDR data transferred to four unseen cities. The model achieves \( R^2 \) values of 0.69, 0.61, 0.48, and 0.42 in LA, SF Bay, Bogot\'a, and Rio de Janeiro. Remarkably, for LA and SF Bay, these zero-shot transfer results closely parallel models trained with a 10\% local observation ratio (\(R^2\) = 0.70 and 0.60). 

NeuroGravity maintains this robust transferability across various data sources, including CDR, LBS, and census commuting flows (Supplementary Figs. 18–19; Supplementary Tables 12–18).
However, baseline models struggle with cross-city generalization. While the gravity model resembles the coarse flow distributions, its two-parameter formulation constrains its pairwise predictive accuracy. DG++ achieves moderate performance in cities with shared socioeconomic profiles but fails to maintain reliable performance across regions with divergent urban morphologies or infrastructural heterogeneities. GNNs, though adept at capturing peripheral flow structures, fail to resolve critical details of primary flow corridors. 
On average, neuroGravity nearly doubles (+99\%) the \( R^2 \) on the CDR data tests (Supplementary Tables~12-13) compared to the best-performing baseline, underscoring its adaptability to cities across diverse geographic, economic, and cultural contexts. Additional analysis, including the distance-binned error, flow ranking error assessments, and per origin/destination marginal error profiles, is detailed in Supplementary Section 7.2-7.4.

We further study the factors impacting the performance of transferring neuroGravity. Fig.~5b visualizes the transfer \( R^2 \)  between five cities, alongside maps of their income spatial distributions. A clear pattern emerges: cities with more uniform income distributions, such as Boston, LA, and SF Bay, exhibit higher transfer \( R^2 \) values when paired. Conversely, cities with pronounced income imbalances, such as Bogot\'a and Rio de Janeiro, face greater challenges in cross-city flow generation. This suggests that spatial income distributions profoundly influence mobility patterns, making flow transfer between income-distribution disparate cities more complex and less accurate.

To quantify this influence, we introduce a spatial income segregation index \( SI \), which measures the degree of income clustering across regions within a city. Unlike traditional spatial segregation metrics such as the dissimilarity index, the rank-order theory index, and the residential income segregation index~\cite{reardon2011measures, reardon2011income, bischoff2014residential}, \( SI \) accounts for spatial characteristics by leveraging the decomposability properties of Bregman Information ($BI$)~\cite{bregman1967relaxation, dhillon2003divisive, chodrow2017structure}. We provide its definition in the Methods section and additional details in Supplementary Section~11. As illustrated in Fig.~5c, we cluster the city's original divisions by merging adjacent areas with relatively similar income levels. The $BI$ decomposition then allows us to separate global income segregation into inter-region and intra-region components. The spatial income segregation index \( SI \) is then defined as the proportion of inter-region segregation to the total global segregation, enabling it to capture the extent of spatial clustering in income distribution. \( SI \) ranges theoretically from 0 to 1. At \(SI = 0\), complete random mixing occurs: microscale disparities vanish entirely when aggregated to broader spatial contexts, reflecting a complete absence of spatial clustering. At the opposing extreme \(SI = 1\), perfect segregation emerges: all disparities are strictly confined to boundaries between clusters, while absolute income equality prevails within each cluster. Finally, cities with higher \( SI \) values exhibit more pronounced spatial income segregation, as detailed in Supplementary Section~11 and Supplementary Fig.~39.

The impact of $SI$ on mobility flow transferability is evident in Fig.~5d. Across experiments where each city serves as the source, transfer \( R^2 \) decreases as the $SI$ disparity between source and target cities increases. This trend suggests that cities with more similar income distributions facilitate better transferability. Furthermore, source cities with lower $SI$ values tend to yield more generalizable models, such as Boston, as their less segregated mobility patterns are more compatible with diverse urban environments. By incorporating the \( SI \) index alongside other factors, such as the area distribution of administrative divisions and the density of OSM features, we performed a multivariate linear regression on the transfer \( R^2 \) across various city pairs, achieving a fitting \( R^2 \) of 0.97 (Fig.~5e). The absolute regression coefficients of the eight most influential factors, presented in Fig.~5f, reveal that the \( SI \) difference is the most significant variable, followed by the source city’s \( SI \) and the target city's \( SI \), reinforcing the above findings.

Finally, we apply the neuroGravity model trained on Boston data to reconstruct mobility networks across more than 1,200 cities and regions worldwide (Fig.~1d). Crucially, we validate neuroGravity's flow estimations against travel surveys in two challenging sub-Saharan African regions (Supplementary~Figs.~25-28), showing strong alignment with actual mobility patterns. This highlights the model's potential as a practical tool for providing mobility insights in data-scarce regions. Furthermore, we adopt neuroGravity-generated flows to drive susceptible-exposed-infectious-recovered (SEIR) pandemic simulations~\cite{chang2021mobility, lai2020effect} for LA and SF Bay. As demonstrated in Supplementary Fig.~36, simulations based on neuroGravity flow estimates closely align with those using ground-truth flows, capturing key epidemiological trends. These applications highlight neuroGravity's potential to inform pandemic analysis efforts in regions with limited availability of mobility data.

\section*{Discussion}

NeuroGravity's physics-informed framework bridges data gaps in urban science, revealing socioeconomic segregation as a fundamental driver of mobility. However, fully realizing this global potential remains constrained by our model's reliance on open-source geographic information.
Recent studies have shown that OSM data exhibits significant spatial biases, with data completeness varying greatly across regions~\cite{milojevic2023eubucco, herfort2023spatio}. In some areas, OSM data covers over 80\% of buildings, while in others, it may cover less than 20\%~\cite{herfort2023spatio}. This uneven coverage can introduce biases into the flow estimation, potentially limiting the model's robustness in cities with incomplete OSM data. In this work, neuroGravity is proactively designed with feature normalization and dropout mechanisms to mitigate this impact. Our empirical tests show that neuroGravity remains stable up to a 30\% OSM feature missing ratio (Supplementary Figs.~30-33).

However, we acknowledge that for the hardest cases in the Global South, the missing rate can fall into the extreme tail of the distribution, exceeding this threshold. In such highly data-scarce scenarios, the data quality remains a persistent challenge, and improving the quality and completeness of the built-environment features in OSM could further enhance the performance of neuroGravity.

Looking forward, there are several promising avenues for future research. First, while our current straightforward concatenation of Google AlphaEarth satellite imagery embeddings~\cite{brown2025alphaearth} yielded marginal gains  (Supplementary Section 8), we recognize that this simple integration method may be suboptimal. There remains potential in developing more sophisticated multimodal fusion strategies and task-specific visual encoders that better capture urban semantics relevant to human movement. Second, the complex interplay between urban structure, mobility behavior, and socioeconomic attributes requires deeper investigation. Our study revealed that neuroGravity embeddings enhance socioeconomic prediction in monocentric cities like Boston, but show diminishing returns in polycentric regions like LA and SF Bay. Future work could aim to unravel the mechanisms behind this heterogeneity.
Lastly, the spatial income segregation index proposed in this study opens possibilities for large-scale sociodemographic analysis. Leveraging this metric, future studies could comprehensively assess the income spatial segregation across the United States and other nations with available data and analyze its correlation with broader determinants to inform more equitable urban planning policies.

Overall, neuroGravity advances urban mobility modeling by leveraging machine learning techniques within a physically informed framework, providing a versatile tool for both flow estimation and socioeconomic analysis in urban systems.

\section*{Methods}
\subsection*{Data description}
\noindent {\bf Built environment data.} 
The 52 distinct built-environment features, used as inputs for the connection predictor and flow estimator in neuroGravity, are derived from OpenStreetMap~\cite{OpenStreetMap}. These features include building, land-use, point-of-interest (POI), and road network characteristics. Building-related features capture the proportion of area occupied by specific building types (e.g., residential and commercial) within a region. Land-use features represent the proportion of area allocated to different land uses (e.g., residential and industrial). POI-related features reflect the average number of specific POIs (e.g., restaurants and schools) per square kilometer. Road network features measure the average road length per square kilometer, indicating the density of transportation infrastructure. These features are critical for estimating mobility flows.

\noindent {\bf Administrative division boundary and population data.} 
For U.S. cities, administrative boundaries are based on ZIP code tabulation areas (ZCTAs) from the Census Bureau~\cite{USCensus2021}. For cities in other countries, administrative divisions vary (Supplementary Table~1), and data is sourced from global databases like GADM~\cite{GADM2024} and Humanitarian Data Exchange (HDX)~\cite{HDX2024}. Population data is obtained and projected onto the regional level, with U.S. population estimates from the Census Bureau and global population estimates from WorldPop~\cite{Worldpop}.

\noindent {\bf Mobility data.} 
The mobility data in this study is derived from two primary sources: CDR and LBS data. CDR data captures the location of mobile phone users when they interact with cellular base stations, while LBS data, typically obtained from mobile applications, provides more precise GPS-based location information. From these data, we extract stay points and construct user trajectories. These datasets correspond to distinct user groups and do not represent the travel behaviors of all residents within a city. To scale these trajectories to the entire population, we use the TimeGeo model~\cite{jiang2016timegeo} for individual travel simulation (Supplementary Section 1.5), followed by aggregation to compute the average daily mobility flow between regions.

\noindent {\bf Socioeconomic and livability data.} 
Socioeconomic indicators for U.S. cities, including household income and educational attainment, are obtained from the U.S. Census 2021 5-year ACS at the ZCTA level. Household carbon footprint data is sourced from CoolClimate~\cite{jones2014spatial}. Nitrogen dioxide (NO$_2$) concentration is mapped to ZCTAs using data from ref.~\cite{wang2023disparities}. The radius of gyration ($Rg$)~\cite{xu2023urban}, a measure of travel range, is calculated based on CDR data as described in ref.~\cite{xu2023urban}. The income data of Rio de Janeiro and Bogot\'a are respectively acquired from ref.~\cite{florez2017measuring,censobr2023}.


\subsection*{Model architecture of neuroGravity}

The pipeline begins by preparing high-dimensional features for each region, represented as $h^0$ (Fig. 2a). These features include population from census data or worldpop~\cite{Worldpop}, built-environment information from the publicly available OpenStreetMap (OSM)~\cite{OpenStreetMap}, which contains land use, natural features (green fields, water areas, beaches) and building type proportions, POI density, and road density (Supplementary Section 3.1). Next, given the natural sparsity of mobility networks~\cite{schlapfer2014scaling}, our plan is to identify the origin-destination (OD) pairs that potentially have mobility flows. As illustrated in Fig.~2b, we employ a LightGBM~\cite{ke2017lightgbm} based binary classifier to predict the presence of stable flows between OD pairs based on their built-environment features and distances, establishing the network structure $\mathcal{E}$ for the graph learning process. {The performance of this connection predictor, validated across multiple cities, is detailed in Supplementary Figs. 4–6 and Supplementary Table 3.}

The third stage is to estimate the flow on each edge in $\mathcal{E}$ with the neuroGravity model. To this end, we propose a deep-parameterized gravity model, referred to as ``meta-Gravity'', to integrate the gravity law into neuroGravity (Fig.~2c). With meta-Gravity, we can generate preliminary flow estimates for each OD pair in $\mathcal{E}$, $[\hat{F}^{g}_{ij}]$.
These estimates, combined with the distance $D_{ij}$ between regions $i$ and $j$, compose the initial edge features for the GNN: $e_{ij}^{(0)} = [D_{ij}, \hat{F}^{g}_{ij}]^\top$. Unlike the traditional gravity model, which uses fixed parameters, neuroGravity employs multi-layer perceptrons (MLPs) to dynamically estimate the gravitational constant $G$ and the distance decay exponent $\alpha$ based on the concatenation of regional features ($h^0_i$ and $h^0_j$). The meta-Gravity model is formulated as:
\begin{equation}
\hat{F}^{g}_{ij} = \frac{G(h^0_i \oplus h^0_j) P_i P_j}{D_{ij}^{\alpha(h^0_i \oplus h^0_j)}},
\end{equation}
where $P_i$ and $P_j$ are the populations of regions $i$ and $j$, and $G(h^0_i \oplus h^0_j)$ and $\alpha(h^0_i \oplus h^0_j)$ are MLP-estimated parameters. 
The meta-Gravity model serves as a low-fidelity network, generating a physically-grounded base flow estimation that encapsulates core mobility principles (distance decay, population attraction). This estimation is incorporated as an initial edge feature, which evolves through the GNN layers and is subsequently concatenated with learned node features for the final log-flow prediction. Within this framework, aligned with the hybrid augmentation paradigm in physics-informed deep learning (PIDL)~\cite{karniadakis2021physics}, the GNN is dedicated to learning the complex, context-specific variations from the physical baseline. Additionally, this initial estimation participates in the attention mechanism during message passing, guiding the information propagation to align with real-world mobility dynamics.

NeuroGravity then employs an edge-enhanced multi-layer graph transformer~\cite{yun2019graph, zhang2020graph} to encode built-environment information into node embeddings $h_j$ and edge embeddings $e_{ij}$, effectively capturing key mobility characteristics. (Fig.~2d).
For the final flow estimation on each OD pair, neuroGravity employs an MLP that integrates the population with the final node and edge embeddings derived from the GNN as input.

Here, we transform the estimation of mobility flow into a logarithmic space, inspired by two key advantages. First, the gravity law posits a multiplicative relationship between the connectivity strengths of two locations, which standard MLPs struggle to model due to their additive nature~\cite{jayakumar2020multiplicative}. The logarithmic transformation linearizes these multiplicative and power relationships into weighted sums, making them more amenable to MLPs with ReLU activations. Second, urban mobility flows typically follow a power-law distribution~\cite{simini2012universal}, characterized by a long tail and significant imbalance across flow magnitudes. Transforming flows into logarithmic space normalizes this distribution into a more balanced linear form (Supplementary Fig.~2), facilitating the model's ability to learn patterns across different flow scales and enhancing convergence during training.
The final flow estimator in logarithmic space is formed as:
\begin{equation}
\log \hat{F}_{ij} = \text{MLP}(P_i, P_j, h_i, h_j, e_{ij}) - \alpha \log D_{ij} + \epsilon,
\end{equation}
where $\epsilon$ and $\alpha$ are learnable scalars corresponding to the logarithmic gravity constant and distance decay exponent, aligning with the traditional gravity model. 

By operating in logarithmic space, neuroGravity’s flow predictor inherently preserves the multiplicative and power-law relationships fundamental to classical gravity laws, integrating their mathematical formalism as a constrained architectural component within the neural network (Supplementary Section~3.5). {This design embodies the inductive bias paradigm in PIDL, as it architecturally encodes the mathematical formalism of physics, providing a strong structural prior.} A Huber-based reconstruction loss is applied across the meta-Gravity and GNN components, enabling end-to-end training that harmonizes physical modeling with graph-based learning.

\subsection*{GNN formulation and loss design}
\noindent{\bf Edge-enhanced Graph-BERT}
The graph transformer in neuroGravity updates node and edge embeddings iteratively across layers using the following equations:
\begin{equation}
\left\{ 
\begin{aligned}
h_j^{(l+1)} &= \gamma^l(h_j^{(l)}) + \sum_{i: (i, j)\in \mathcal{E}} \Lambda^{l}(h_j^{(l)}, h_i^{(l)}, e_{ij}^{(l)}) \, m^l(h_i^{(l)}, e_{ij}^{(l)}) \\
e_{ij}^{(l+1)} &= \psi^l (e_{ij}^{(l)}, h_i^{(l+1)}, h_j^{(l+1)}) \\
\end{aligned}
\right.,
\end{equation}
where $h_j^{(l)}$ and $e_{ij}^{(l)}$ respectively denote the node embedding of region $i$ and edge embedding between region $i$ and $j$ at GNN layer $l$. $\gamma^l(\cdot)$, $m^l(\cdot)$, and $\psi^l(\cdot)$ respectively correspond to the MLP-based projection functions for node residuals, link message, and edge feature, the details are elaborated in Supplementary Section~3.

The edge-regulated attention weight $\Lambda^l(h_i^{(l)}, h_j^{(l)}, e_{ij}^{(l)})$ modulates the message passing based on edge features, aligning message flow with real-world mobility dynamics. It is calculated as:
\begin{equation}
\Lambda^l(h_i^{(l)}, h_j^{(l)}, e_{ij}^{(l)}) = \text{Softmax}_{i: (i, j) \in \mathcal{E}}\left(\frac{W_Q^{l}h_j^{(l)}\cdot(W_K^{l}h_i^{(l)} + W_{KE}^{l}e_{ij}^{(l)})}{\sqrt{d_\text{qkv}^{(l)}}}\right),
\end{equation}
where \(W_Q^{l}, W_K^{l}\in \mathbb{R}^{d_\text{qkv}^{(l)} \times d_\text{node}^{(l)}}, W_{KE}^{l} \in \mathbb{R}^{d_\text{qkv}^{(l)} \times d_\text{edge}^{(l)}}\) are the learnable projection matrices, \(d_\text{qkv}^{(l)}\) is the dimensionality of the query and key vectors, and the subscript \(i: (i, j) \in \mathcal{E}\) indicates that the Softmax normalization is performed over all source nodes targeting node $j$. For brevity, the multi-head mechanism is not explicitly represented here; in practice, this attention process is carried out across multiple heads, with their resulting messages concatenated to capture diverse aspects of the relationships between nodes and edges.

\noindent{\bf Training strategy and loss function}
To optimize the flow estimation, we design two Huber-based loss functions to train the two pipelines of neuroGravity:

\begin{equation}
    \mathcal{L}^g = \sum_{(i,j) \in \mathcal{E}_{\text{obs}}} w_{ij} \cdot \mathcal{Q}_{\delta}( \log \hat{F}_{ij}^g - \log F_{ij})
\end{equation}
\begin{equation}
    \mathcal{L} = \sum_{(i,j) \in \mathcal{E}_{\text{obs}}} w_{ij} \cdot \mathcal{Q}_{\delta}( \log \hat{F}_{ij} - \log F_{ij} )
\end{equation}

\normalcolor

Specifically, \(\mathcal{L}^g\) is utilized to pretrain the physics-model pipeline first, and then $\mathcal{L}$ is employed to train both pipelines jointly.
Here, \(\hat{F}_{ij}\) and \(\hat{F}_{ij}^g\) respectively are the flow estimations from the GNN pipeline and the meta-Gravity pipeline, 
and \(\mathcal{E}_{\text{obs}}\) is the set of observed region pairs with flow greater than zero. 
{The Huber loss function \(\mathcal{Q}_{\delta}\) is formulated as follows, with the threshold parameter \(\delta\) controlling the transition point from quadratic to linear set to be 0.5.
\begin{equation}
\mathcal{Q}_{\delta}(r) =
\begin{cases}
\frac{1}{2} r^2, & \text{if } |r| \leq \delta, \\
\delta (|r| - \frac{1}{2} \delta), & \text{otherwise},
\end{cases}
\end{equation}

The assignment of sample weights \(w_{ij}\) differs between tasks due to the amount of available training data. In the cross-city transfer task, where the training samples are sufficient, we set \(w_{ij} = 1\) for all observed flows. The model leverages abundant training data to automatically distinguish dominant mobility patterns from potential observation noise and systematic biases. Conversely, in the few-shot reconstruction task where training samples are extremely scarce, we employ a weighted scheme:
\begin{equation}
w_{ij} = \frac{\exp(F_{ij}/\tau)}{\sum_{(k,l) \in \mathcal{E}_{\text{obs}}} \exp(F_{kl}/\tau)},
\end{equation}
with a temperature parameter \(\tau = 2\). This formulation prioritizes flows with higher magnitudes, guiding the model to focus on more trustworthy observations when training samples are scarce. 

An early stopping criterion was implemented to prevent overfitting. Internal flows from 20\% of the observed zones were further reserved as a validation set. Training was stopped if neither the R\(^2\) nor CPC metric improved for 100 consecutive epochs on the validation set, after which the model parameters were restored to the checkpoint with the highest validation R\(^2\).
}

\subsection*{Evaluation metrics}
To quantitatively assess flow estimation accuracy, we employ the $R^2$ and the common part of commuters (CPC)\cite{lenormand2016systematic}:
\begin{equation}
R^2 = 1 - \frac{\sum (F_{ij} - \hat{F}_{ij})^2}{\sum (F_{ij} - \overline{F})^2} 
\qquad \text{and} \qquad 
\text{CPC} = \frac{2 \sum \min(F_{ij}, \hat{F}_{ij})}{\sum F_{ij} + \sum \hat{F}_{ij}},
\end{equation}
where $F_{ij}$ represents the actual flows, $\hat{F}_{ij}$ the estimated flows, and $\overline{F}$ the mean of actual flows. The $R^2$ evaluates the proportion of variance captured by the model, while the CPC metric measures the overlap between the observed and estimated flow distributions (Supplementary Fig. 3). For a more comprehensive evaluation, additional ranking-based metrics including Spearman’s Rank Correlation Coefficient, Recall@K, and Normalized Discounted Cumulative Gain at K (nDCG@K) and the corresponding model performances are detailed in Supplementary Section 7.3.

\subsection*{Spatial income segregation index}
The overall income segregation of the city is defined using the Bregman information~\cite{banerjee2005clustering}, which is the population-weighted average of the local divergences from the global income:
\begin{equation}
I(\mathcal{R}, Y) = \sum_{r_i \in \mathcal{R}} \frac{P_i}{\sum_{r_j \in \mathcal{R}} P_j} \, \left[y_i^2 - \bar{y}^2 - 2(y_i - \bar{y})\right],
\end{equation}
where \( \mathcal{R} \) denotes the set of all regions in the city, \( Y = \{ y_i \} \) is the set of local average incomes, \( P_i \) is the population of region \( r_i \) and \(\bar{y} = \frac{\sum_{r \in \mathcal{R}} P_r y_r}{\sum_{r \in \mathcal{R}} P_r} \) is the average income of region \( \mathcal{R} \).

Iteratively aggregate adjacent regions with similar income, we can get $K$ larger divisions $\mathcal{R}^K=\{\mathcal{R}^K_k\subseteq\mathcal{R}\mid k=1, ..., K\}$ with their corresponding income set $Y^K=\{Y^K_k\subseteq{Y}\mid k=1, ..., K\}$. By the chain rule of Bregman information~\cite{dhillon2003divisive}, the total income segregation can be decomposed into inter-region segregation and intra-region segregation on the $K$ aggregated divisions:
\begin{equation}
I(\mathcal{R}, Y) = \underbrace{I(\mathcal{R}^K, Y^K)}_{BI_\text{inter}} + \underbrace{\sum_{k = 1}^K \frac{\sum_{r_i \in \mathcal{R}_{k}^K} P_i}{\sum_{r_j \in \mathcal{R}} P_j} \, I(\mathcal{R}_{k}^K, Y_{k}^K)}_{BI_\text{intra}}.
\end{equation}
The spatial income segregation index \(SI\) is then defined as the fraction of inter-region segregation to the total segregation. The details are elaborated in Supplementary Section~11.
\begin{equation}
{SI} = \frac{I(\mathcal{R}^K, Y^K)}{I(\mathcal{R}, Y)} = \frac{{BI}_{\text{inter}}}{{BI}_{\text{inter}} + {BI}_{\text{intra}}}.
\end{equation}

\section*{Data availability} 
All data needed to evaluate the conclusions in the paper are described in the paper and the Supplementary Information. For contractual and privacy reasons, we cannot make the raw mobile-phone data available. One can contract Kido Dynamics SA to try to get access to the raw mobile-phone data.
Public datasets for commuting flows in the United Kingdom and Italy, analyzed in the Supplementary Information, are available from the UK Data Service (\url{https://wicid.ukdataservice.ac.uk/cider/wicid/downloads.php}) and ISTAT (\url{https://www.istat.it/non-categorizzato/matrici-del-pendolarismo/}). Source data for Figures 3, 4, and 5 are provided with this paper. The neuroGravity estimated mobility networks are available in~\cite{ng_mobility_networks}.

\section*{Code availability}
The implementation of this work is available at GitHub~(\url{https://github.com/urbanmobility/NeuroGravity}) and Zenodo~\cite{ng}.

\section*{Acknowledgments}
We thank the SJTU AI for Science platform for computing support. This work was jointly supported by the Shanghai Municipal Science and Technology Major Project (2021SHZDZX0102), the National Natural Science Foundation of China (72432007), and the Fundamental Research Funds for the Central Universities.

\section*{Author contributions}
YX and JY conceived the research and designed the model and analyses. MCG provided the mobile phone data and discussed the analyses. JY, SH and ZH processed the mobile phone data. JY performed the analyses. YJ and XY built the computing platform for model training. JY, YX and MCG wrote the paper. YX, YJ and XY supervised the research.

\section*{Competing interests}
The authors declare that they have no competing interests.

\section*{Inclusion \& ethics}
All contributors fulfill the authorship criteria are listed as co-authors in this paper. Other contributors who do not meet all criteria for authorship are listed in the Acknowledgements.



\clearpage

\begin{figure*}[htb!]
\centering
\includegraphics[width=0.95\linewidth]{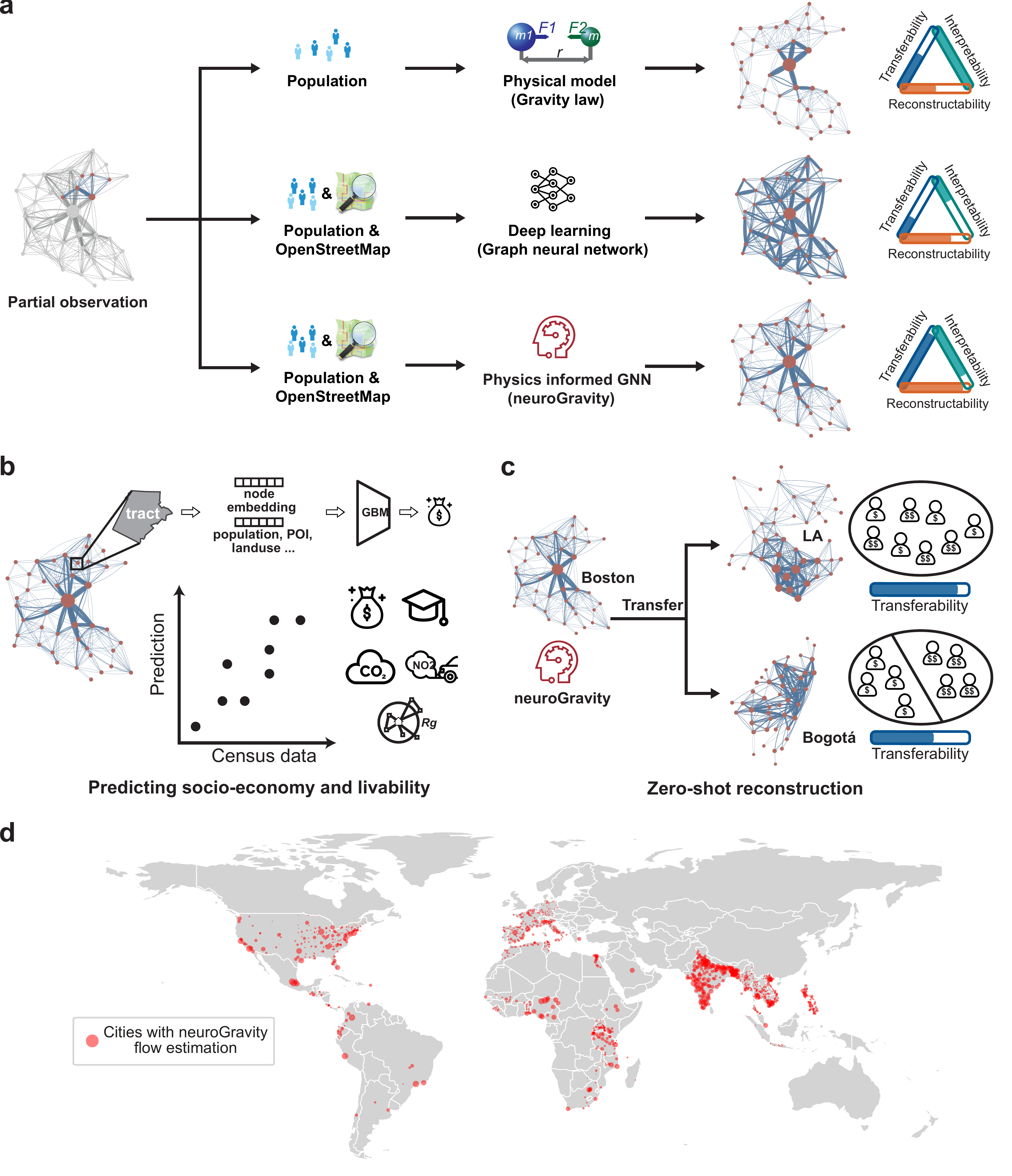}
\end{figure*}
\noindent{{\bf Fig. 1. Conception and potential capabilities of neuroGravity.} (\textbf{a}) Reconstruction of city-wide mobility flows based on sparse partial observations, comparing three approaches: the gravity-based physical model, which is highly interpretable but lacks sufficient data-fitting capability; the GNN model, which is prone to overfitting to observed flow patterns and can generate unrealistic predictions in data-scarce regions, compromising its adaptability; and neuroGravity, a physics-informed GNN that integrates physical insights with deep graph learning, demonstrating enhanced reconstructability, transferability, and interpretability. (\textbf{b}) Inference of fine-scale socioeconomic and livability statistics using OSM data and neuroGravity node embeddings. GBM is utilized to infer these statistics, including household income, the proportion of the population with a bachelor’s degree, household carbon footprint, Nitrogen Dioxide concentration, and the radius of gyration. (\textbf{c}) Zero-shot reconstruction of mobility flows in unobserved cities. We transfer the well-trained neuroGravity models from one city to the others. Numerical validations reveal that transferability is significantly influenced by the income spatial segregation of the source and the target city.} (\textbf{d}) NeuroGravity has been applied to estimate mobility flows across over 1,200 cities and regions worldwide~\cite{ng_mobility_networks}, providing flow proxies for mobility-powered tasks and studies in regions where ground-truth observations are scarce or unavailable.

\newpage

\begin{figure*}[tb!]
\centering
\includegraphics[width=1.0\linewidth]{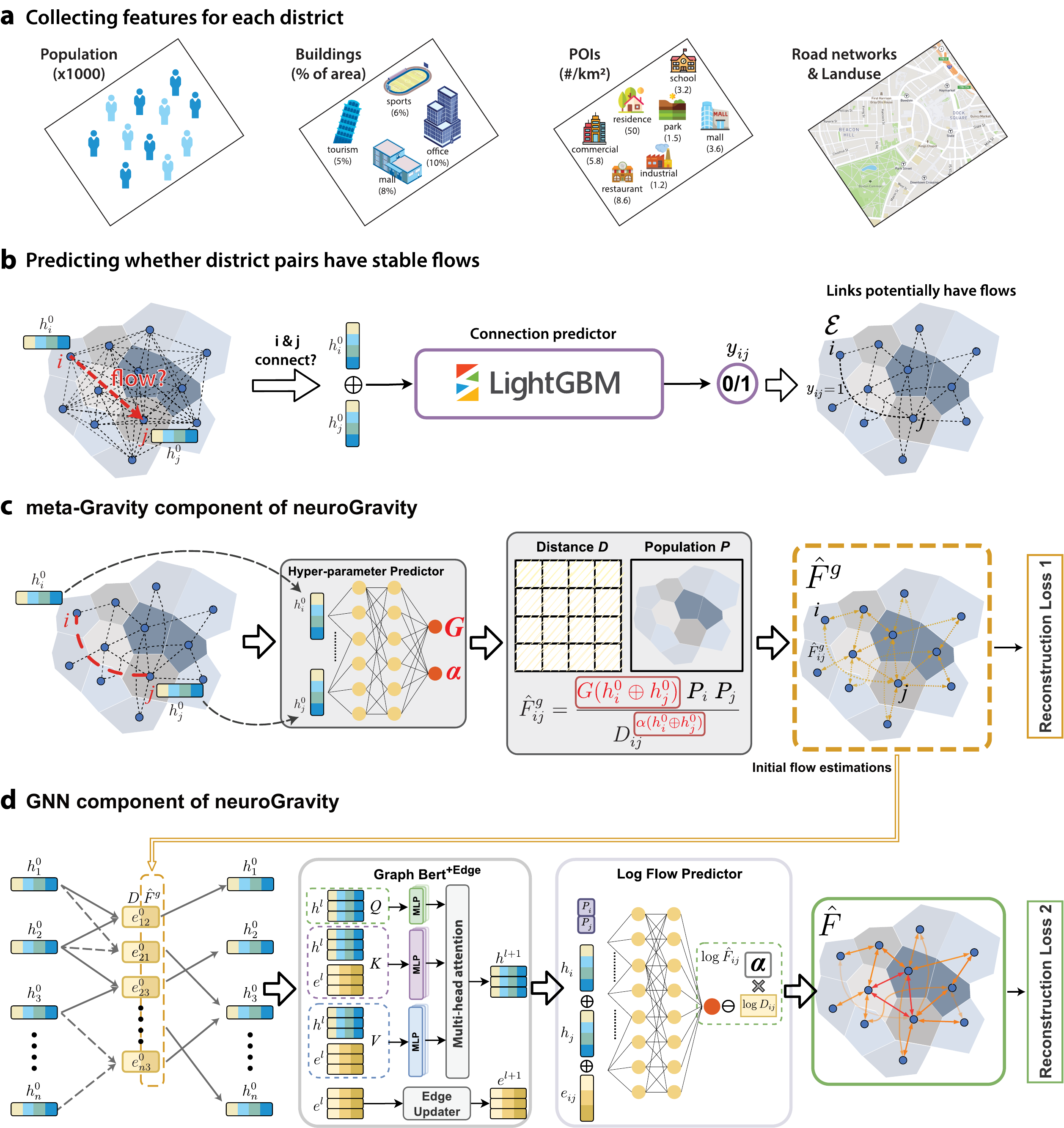}
\end{figure*}
\noindent{{\bf Fig. 2. Schematic illustration of the neuroGravity model architecture.} (\textbf{a}) Feature preparation for each district, including population, area percentages of various building types, the number of different POIs per square kilometer, area percentages of different land uses, and road lengths of various classes per square kilometer. (\textbf{b}) Identification of district pairs with potential mobility flows using the LightGBM model. Driven by the initial node features (\(h_i^0\) and \(h_j^0\)) of districts \( i \) and \( j \), the model predicts a binary indicator (\( y_{ij} \in \{0, 1\} \)) to infer flow existence, forming the set of potentially connected edges (\( \mathcal{E} \)). GBM is trained with the origin-destination (OD) pairs within the observed regions \( \mathcal{R}_{\text{obs}} \). (\textbf{c}) The gravity model is deep-parameterized as meta-Gravity. It estimates the gravitational parameter (\(G\)) and the distance decay exponent (\(\alpha\)) using MLPs based on the concatenated (\(\oplus\)) features of the OD pairs. Using populations (\(P_i, P_j\)) and spatial distance (\(D_{ij}\)), it generates an initial flow estimation (\(\hat{F}^g_{ij}\)). (\textbf{d}) Final flow reconstruction using an edge-enhanced Graph-BERT model. Initial edge features (\(e_{ij}^0\)) are initialized with \(D_{ij}\) and \(\hat{F}^g_{ij}\). Node and edge embeddings at layer \(l\) (\(h^l, e^l\)) are updated via Query (\(Q\)), Key (\(K\)), and Value (\(V\)) multi-head attention mechanisms to produce next-layer representations. To act in a gravity-like manner, the Log Flow Predictor utilizes these final embeddings alongside population and distance data to estimate the flow in the logarithmic space (\(\log \hat{F}_{ij}\)), which then yields the final flow (\(\hat{F}_{ij}\)). The components in (\textbf{c-d}) are jointly optimized by a Huber loss through back-propagation.}

\newpage

\begin{figure*}[tb!]
\centering
\includegraphics[width=1.0\linewidth]{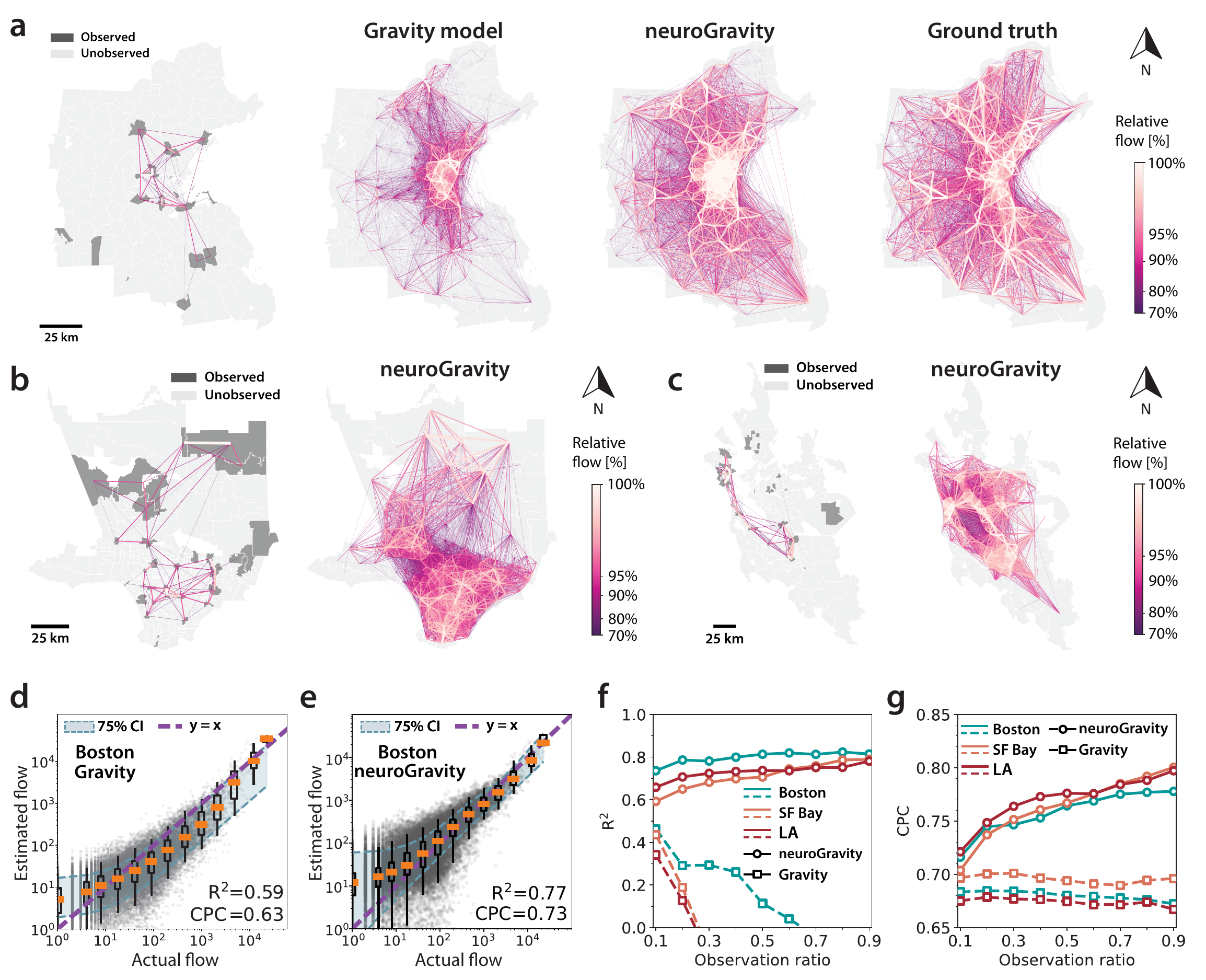}
\end{figure*}
\noindent{{\bf Fig. 3. Performance evaluation of neuroGravity for mobility flow reconstruction with partial observations.} (\textbf{a}) Visualization of the observed flows, the reconstructed flows using the Gravity and neuroGravity models, and the ground truth in Boston. The observed flows, serving as the training set, are the internal travels under 10\% observation ratio, accounting for 1\% \ of the valid flow links. (\textbf{b-c}) Visualization of observed flows and the reconstructed flows via neuroGravity in Los Angeles (\( R^2 \) = 0.70) and San Francisco Bay area (\( R^2 \) = 0.60). (\textbf{d-e}) Estimated versus actual flows in Boston under the 10\% observation ratio, comparing the gravity model (\( R^2 \) = 0.59) and neuroGravity (\( R^2 \) = 0.77). Each point signifies an individual flow (\(n=51,786\) origin-destination pairs), and denser clustering along the \( y = x \) line reflects higher accuracy. For the box plots, the central line indicates the median, the bounds of the box represent the 25th and 75th percentiles, and the whiskers denote the minima and maxima values within 1.5 $\times$ the interquartile range (IQR). The shaded regions denote the 75\% confidence intervals (12.5th to 87.5th percentiles smoothed via locally estimated scatterplot smoothing (LOESS)), centered on the smoothed median. (\textbf{f-g}) Median \( R^2 \) and CPC values of flow reconstructions across varying observation ratios and cities, comparing the Gravity and neuroGravity models. Each data point derives from \( n = 30 \) independent experiments with randomly sampled observations.}

\newpage

\begin{figure*}[tb!]
\centering
\includegraphics[width=1.0\linewidth]{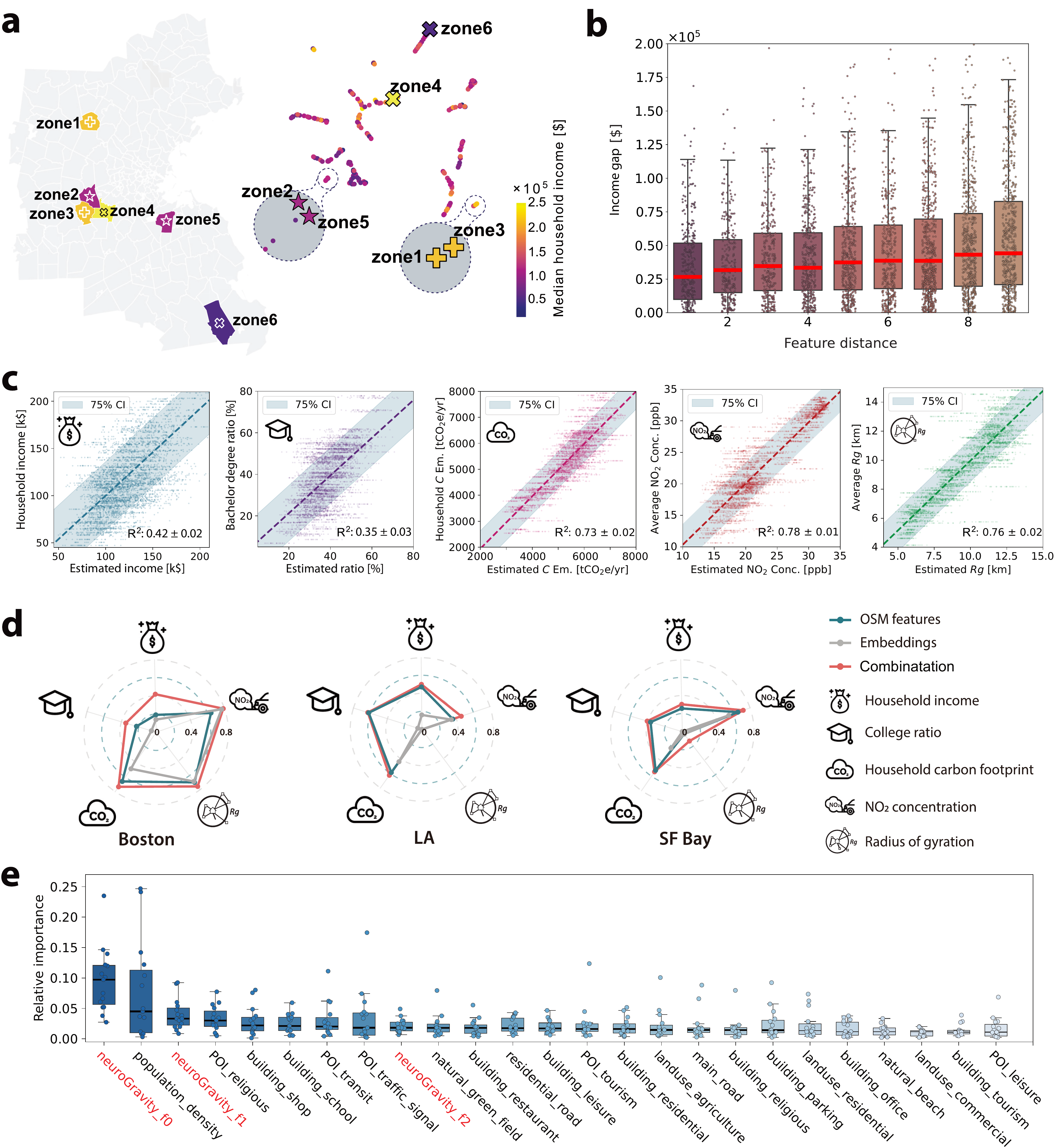}
\end{figure*}
\noindent{{\bf Fig. 4. Inference of socioeconomic and livability statistics using neuroGravity node embeddings.} (\textbf{a}) 2-dimensional UMAP projections of neuroGravity node embeddings for Boston ZIP code areas (ZCTAs), colored by median household income. Despite being untrained on income data, embeddings capture socioeconomic patterns, clustering geographically distant but economically similar ZCTAs (e.g., zone2 vs. zone5 and zone1 vs. zone3).  (\textbf{b}) Pair-wise household income gaps between two districts show a monotonically increasing trend with the increase of embedding distances (\(n = 62,250\) ZCTA pairs). For the box plots, the red central lines denote the median income gap for each distance bin, the bounds of the boxes represent the 25th and 75th percentiles, and the whiskers extend to the minimum and maximum values within 1.5 $\times$ the interquartile range (IQR).
(\textbf{c}) Estimation performance for five attributes in Boston using GBMs trained on combined embeddings and OSM features. Scatter plots show aggregated 20-round 5-fold cross-validation results. Dashed lines represent mean Ordinary Least Squares (OLS) predictions, and shaded regions indicate 75\% confidence intervals. Overall performance \(R^2\) is labeled for each attribute as mean $\pm$ standard deviation across the bootstrap iterations.
(\textbf{d}) Radar charts comparing the average \(R^2\) from 20 rounds of bootstrap 5-fold CV tests for attribute estimation using OSM features alone, embeddings alone, and their combination across three cities.
(\textbf{e}) SHAP analysis of the relative importance of neuroGravity embeddings and OSM features in estimating the five metrics across three cities. Each box summarizes the SHAP relative importance for a given variable across \(n=15\) independent experiments (3 cities, 5 attributes). Boxes show median, IQR (25th–75th percentile), and whiskers extending to the furthest points within 1.5\(\times\) IQR.}

\newpage

\begin{figure*}[htb!]
\centering
\includegraphics[width=0.95\linewidth]{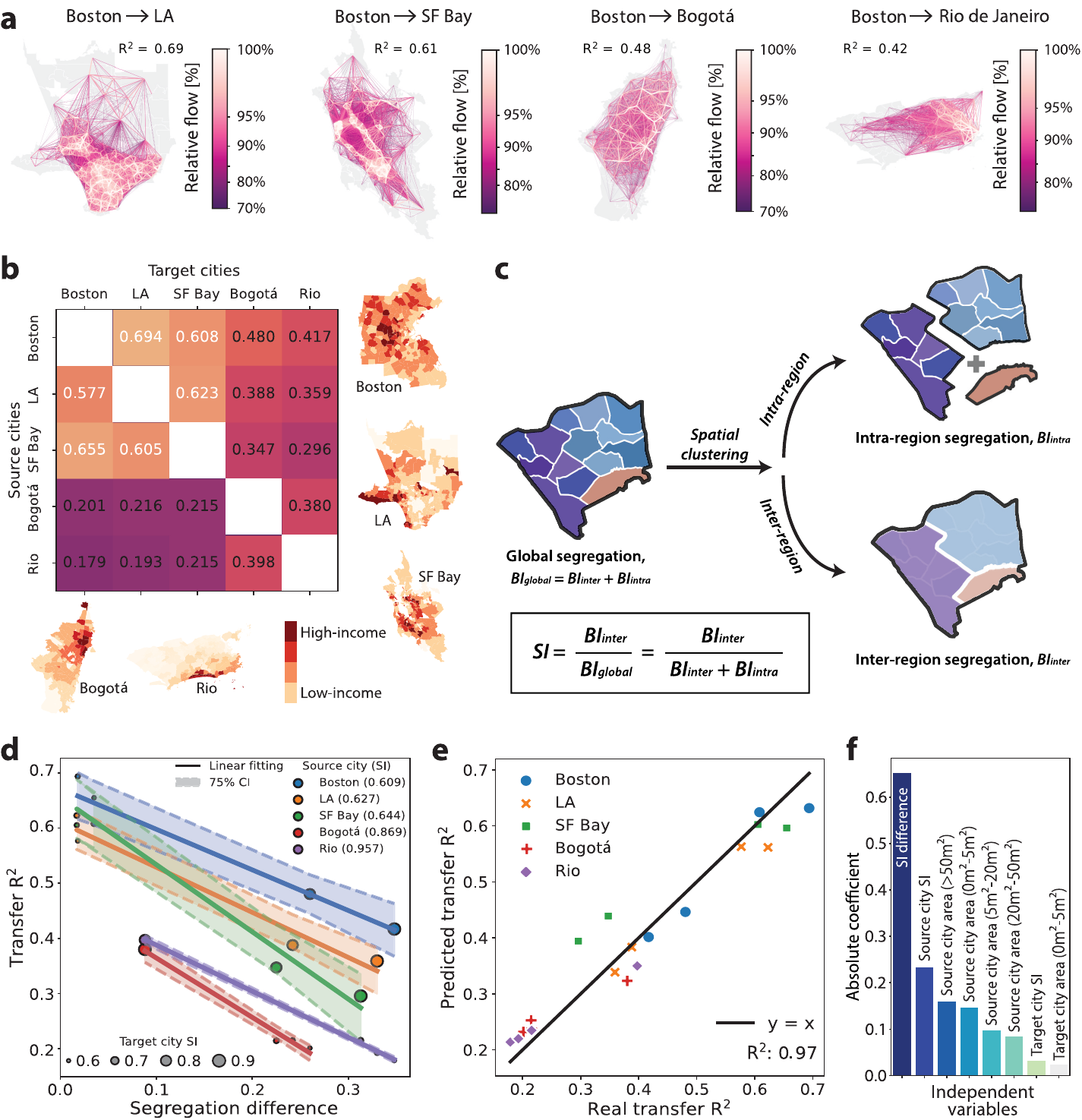}
\end{figure*}
\noindent{{\bf Fig. 5. Assessment of cross-city mobility flow generation performance and its connection to spatial income segregation.} (\textbf{a}) Visualization of the zero-shot mobility flow reconstruction results in Los Angeles, SF Bay, Bogot\'a, and Rio de Janeiro using neuroGravity trained on Boston data. (\textbf{b}) Heatmap of pairwise transfer \( R^2 \) values between cities, illustrating more efficient transfer between cities with uniform income distributions (Boston, LA, SF Bay), and more challenging transfer for cities with relatively high spatial income segregation.
(\textbf{c}) Definition of the spatial segregation index ($SI$) as the ratio of inter-region segregation $BI_{inter}$ to global segregation $BI_{global}$. Higher $SI$ indicates greater spatial income imbalance. (\textbf{d}) Relationship between the transfer \( R^2 \) and the SI difference between source and target cities. Each solid line represents the mean Ordinary Least Squares (OLS) linear fit of transfer \( R^2 \) against $SI$ difference when using a specific source city ($n = 4$ independent cross-city transfer pairs per line). The shaded regions indicate the corresponding 75\% confidence intervals (CI). Greater segregation differences reduce transfer accuracy, as seen in the downward trends of linear fits. Lower segregation in the source city correlates with higher transfer accuracy, as indicated by the relative heights of the five lines. (\textbf{e}) Comparison of real transfer \( R^2 \) with predictions from a linear model using $SI$, region area distribution, and OSM data density. (\textbf{f}) Absolute coefficients of the top 8 independent variables in the linear regression for transfer \( R^2 \), highlighting segregation difference as the most significant, followed by segregation of the source city.}

\newpage



\begin{thebibliography}{10}

\bibitem{batty2008size}
M.~Batty, The size, scale, and shape of cities, {\it Science\/} {\bf 319},  769--771 (2008).

\bibitem{bettencourt2013origins}
L.~M. Bettencourt, The origins of scaling in cities, {\it Science\/} {\bf 340},  1438--1441 (2013).

\bibitem{verbavatz2020growth}
V.~Verbavatz, M.~Barthelemy, The growth equation of cities, {\it Nature\/} {\bf 587},  397--401 (2020).

\bibitem{gonzalez2008understanding}
M.~C. Gonzalez, C.~A. Hidalgo, A.-L. Barabasi, Understanding individual human mobility patterns, {\it Nature\/} {\bf 453},  779--782 (2008).

\bibitem{xu2018planning}
Y.~Xu, S.~{\c{C}}olak, E.~C. Kara, S.~J. Moura, M.~C. Gonz{\'a}lez, Planning for electric vehicle needs by coupling charging profiles with urban mobility, {\it Nature Energy\/} {\bf 3},  484--493 (2018).

\bibitem{vazifeh2018addressing}
M.~M. Vazifeh, P.~Santi, G.~Resta, S.~H. Strogatz, C.~Ratti, Addressing the minimum fleet problem in on-demand urban mobility, {\it Nature\/} {\bf 557},  534--538 (2018).

\bibitem{buckee2020aggregated}
C.~O. Buckee, {\it et~al.\/}, Aggregated mobility data could help fight {COVID-19}, {\it Science\/} {\bf 368},  145--146 (2020).

\bibitem{tian2020investigation}
H.~Tian, {\it et~al.\/}, An investigation of transmission control measures during the first 50 days of the {COVID-19} epidemic in {China}, {\it Science\/} {\bf 368},  638--642 (2020).

\bibitem{barbosa2018human}
H.~Barbosa, {\it et~al.\/}, Human mobility: Models and applications, {\it Physics Reports\/} {\bf 734},  1--74 (2018).

\bibitem{CTPP2024}
{American Association of State Highway and Transportation Officials}, Census transportation planning products, {https://transportation.org/ctpp/datasets/} (2016). [Online; accessed 01-June-2024].

\bibitem{toole2015path}
J.~L. Toole, {\it et~al.\/}, The path most traveled: Travel demand estimation using big data resources, {\it Transportation Research Part C: Emerging Technologies\/} {\bf 58},  162--177 (2015).

\bibitem{DigitalEconomy2024}
{United Nations}, Digital economy report 2024, {https://unctad.org/publication/digital-economy-report-2024/} (2024). [Online; accessed 23-September-2024].

\bibitem{moro2021mobility}
E.~Moro, D.~Calacci, X.~Dong, A.~Pentland, Mobility patterns are associated with experienced income segregation in large {US} cities, {\it Nature Communications\/} {\bf 12},  1--10 (2021).

\bibitem{xu2023urban}
Y.~Xu, {\it et~al.\/}, Urban dynamics through the lens of human mobility, {\it Nature Computational Science\/} {\bf 3},  611--620 (2023).

\bibitem{nilforoshan2023human}
H.~Nilforoshan, {\it et~al.\/}, Human mobility networks reveal increased segregation in large cities, {\it Nature\/} {\bf 624},  586--592 (2023).

\bibitem{zipf1946p}
G.~K. Zipf, The {P\textsubscript{1}P\textsubscript{2}/D} hypothesis: on the intercity movement of persons, {\it American Sociological Review\/} {\bf 11},  677--686 (1946).

\bibitem{simini2012universal}
F.~Simini, M.~C. Gonz{\'a}lez, A.~Maritan, A.-L. Barab{\'a}si, A universal model for mobility and migration patterns, {\it Nature\/} {\bf 484},  96--100 (2012).

\bibitem{ganin2017resilience}
A.~A. Ganin, {\it et~al.\/}, Resilience and efficiency in transportation networks, {\it Science Advances\/} {\bf 3},  e1701079 (2017).

\bibitem{kraemer2017spread}
M.~U. Kraemer, {\it et~al.\/}, Spread of yellow fever virus outbreak in angola and the democratic republic of the congo 2015--16: a modelling study, {\it The Lancet infectious diseases\/} {\bf 17},  330--338 (2017).

\bibitem{simini2021deep}
F.~Simini, G.~Barlacchi, M.~Luca, L.~Pappalardo, A deep gravity model for mobility flows generation, {\it Nature communications\/} {\bf 12},  6576 (2021).

\bibitem{song2024enhancing}
R.~Song, G.~Spadon, R.~Pelot, S.~Matwin, A.~Soares, Enhancing global maritime traffic network forecasting with gravity-inspired deep learning models, {\it Scientific Reports\/} {\bf 14},  16665 (2024).

\bibitem{zhao2024predicting}
Y.~Zhao, S.~Cheng, S.~Gao, P.~Wang, F.~Lu, Predicting origin-destination flows by considering heterogeneous mobility patterns, {\it Sustainable Cities and Society\/}  106015 (2024).

\bibitem{cabanas2025human}
O.~Cabanas-Tirapu, L.~Danús, E.~Moro, M.~Sales-Pardo, R.~Guimerà, Human mobility is well described by closed-form gravity-like models learned automatically from data, {\it Nature Communications\/} {\bf 16},  1336 (2025).

\bibitem{chodrow2017structure}
P.~S. Chodrow, Structure and information in spatial segregation, {\it Proceedings of the National Academy of Sciences\/} {\bf 114},  11591--11596 (2017).

\bibitem{chang2021mobility}
S.~Chang, {\it et~al.\/}, Mobility network models of covid-19 explain inequities and inform reopening, {\it Nature\/} {\bf 589},  82--87 (2021).

\bibitem{lai2020effect}
S.~Lai, {\it et~al.\/}, Effect of non-pharmaceutical interventions to contain covid-19 in china, {\it Nature\/} {\bf 585},  410--413 (2020).

\bibitem{lenormand2016systematic}
M.~Lenormand, A.~Bassolas, J.~J. Ramasco, Systematic comparison of trip distribution laws and models, {\it Journal of Transport Geography\/} {\bf 51},  158--169 (2016).

\bibitem{becht2019dimensionality}
E.~Becht, {\it et~al.\/}, Dimensionality reduction for visualizing single-cell data using umap, {\it Nature Biotechnology\/} {\bf 37},  38--44 (2019).

\bibitem{abdi2010principal}
H.~Abdi, L.~J. Williams, Principal component analysis, {\it Wiley Interdisciplinary Reviews: Computational Statistics\/} {\bf 2},  433--459 (2010).

\bibitem{scott2017unified}
M.~Scott, L.~Su-In, {\it et~al.\/}, A unified approach to interpreting model predictions, {\it Advances in Neural Information Processing Systems\/} {\bf 30},  4765--4774 (2017).

\bibitem{reardon2011measures}
S.~F. Reardon, Measures of income segregation, {\it Working paper\/}, Stanford Center for Education Policy Analysis, Stanford, CA (2011).

\bibitem{reardon2011income}
S.~F. Reardon, K.~Bischoff, Income inequality and income segregation, {\it American Journal of Sociology\/} {\bf 116},  1092--1153 (2011).

\bibitem{bischoff2014residential}
K.~Bischoff, S.~F. Reardon, Residential segregation by income, 1970--2009, {\it Diversity and Disparities: America Enters a New Century\/} {\bf 43} (2014).

\bibitem{bregman1967relaxation}
L.~M. Bregman, The relaxation method of finding the common point of convex sets and its application to the solution of problems in convex programming, {\it USSR Computational Mathematics and Mathematical Physics\/} {\bf 7},  200--217 (1967).

\bibitem{dhillon2003divisive}
I.~S. Dhillon, S.~Mallela, R.~Kumar, A divisive information theoretic feature clustering algorithm for text classification, {\it The Journal of Machine Learning Research\/} {\bf 3},  1265--1287 (2003).

\bibitem{milojevic2023eubucco}
N.~Milojevic-Dupont, {\it et~al.\/}, Eubucco v0. 1: European building stock characteristics in a common and open database for 200+ million individual buildings, {\it Scientific Data\/} {\bf 10},  147 (2023).

\bibitem{herfort2023spatio}
B.~Herfort, S.~Lautenbach, J.~Porto~de Albuquerque, J.~Anderson, A.~Zipf, A spatio-temporal analysis investigating completeness and inequalities of global urban building data in openstreetmap, {\it Nature Communications\/} {\bf 14},  3985 (2023).

\bibitem{brown2025alphaearth}
C.~F. Brown, {\it et~al.\/}, Alphaearth foundations: An embedding field model for accurate and efficient global mapping from sparse label data, {\it arXiv preprint arXiv:2507.22291\/}  (2025).

\bibitem{OpenStreetMap}
{OpenStreetMap contributors}, {Planet dump retrieved from https://planet.osm.org }, \url{ https://www.openstreetmap.org } (2017).

\bibitem{USCensus2021}
U.S. Census Bureau, {\it 2020 Census Zip Code Tabulation Areas (ZCTAs)\/} (2021). \url{https://www.census.gov/programs-surveys/geography/guidance/geo-areas/zctas.html}.

\bibitem{GADM2024}
GADM, {\it GADM database of Global Administrative Areas\/} (2023). \url{https://gadm.org}.

\bibitem{HDX2024}
Humanitarian Data Exchange (HDX), United Nations OCHA, {\it Administrative Boundaries and Datasets\/} (2023). \url{https://data.humdata.org}.

\bibitem{Worldpop}
Worldpop: High-resolution population data, WorldPop, School of Geography and Environmental Science, University of Southampton (2024). \url{https://www.worldpop.org}.

\bibitem{jiang2016timegeo}
S.~Jiang, {\it et~al.\/}, The {TimeGeo} modeling framework for urban mobility without travel surveys, {\it Proceedings of the National Academy of Sciences\/} {\bf 113},  E5370--E5378 (2016).

\bibitem{jones2014spatial}
C.~Jones, D.~M. Kammen, Spatial distribution of us household carbon footprints reveals suburbanization undermines greenhouse gas benefits of urban population density, {\it Environmental Science \& Technology\/} {\bf 48},  895--902 (2014).

\bibitem{wang2023disparities}
Y.~Wang, {\it et~al.\/}, Disparities in ambient nitrogen dioxide pollution in the united states, {\it Proceedings of the National Academy of Sciences\/} {\bf 120},  e2208450120 (2023).

\bibitem{florez2017measuring}
M.~A. Florez, {\it et~al.\/}, Measuring the impact of economic well being in commuting networks--a case study of bogota, colombia, {\it Transportation Research Board 96th Annual Meeting\/}, no. 17-03745 (2017).

\bibitem{censobr2023}
{Rafael H. M. Pereira and Rogério J. Barbosa}, {\it {censobr: Download Data from Brazil's Population Census}\/} (2023).

\bibitem{schlapfer2014scaling}
M.~Schl{\"a}pfer, {\it et~al.\/}, The scaling of human interactions with city size, {\it Journal of the Royal Society Interface\/} {\bf 11},  20130789 (2014).

\bibitem{ke2017lightgbm}
G.~Ke, {\it et~al.\/}, Lightgbm: A highly efficient gradient boosting decision tree, {\it Advances in Neural Information Processing Systems\/} (2017), vol.~30,  3149--3157.

\bibitem{karniadakis2021physics}
G.~E. Karniadakis, {\it et~al.\/}, Physics-informed machine learning, {\it Nature Reviews Physics\/} {\bf 3},  422--440 (2021).

\bibitem{yun2019graph}
S.~Yun, M.~Jeong, R.~Kim, J.~Kang, H.~J. Kim, Graph transformer networks, {\it Advances in Neural Information Processing Systems\/} {\bf 32} (2019).

\bibitem{zhang2020graph}
J.~Zhang, H.~Zhang, C.~Xia, L.~Sun, Graph-bert: Only attention is needed for learning graph representations, {\it arXiv preprint arXiv:2001.05140\/}  (2020).

\bibitem{jayakumar2020multiplicative}
S.~M. Jayakumar, {\it et~al.\/}, Multiplicative interactions and where to find them, {\it International Conference on Learning Representations\/} (2020).

\bibitem{banerjee2005clustering}
A.~Banerjee, S.~Merugu, I.~S. Dhillon, J.~Ghosh, J.~Lafferty, Clustering with bregman divergences., {\it Journal of Machine Learning Research\/} {\bf 6} (2005).

\bibitem{ng_mobility_networks}
J.~Yang, {\it et~al.\/}, {NeuroGravity Estimated Mobility Networks for Over 1,200 Cities and Regions Worldwide} (2026). \url{https://doi.org/10.5281/zenodo.19727864}.

\bibitem{ng}
J.~Yang, {\it et~al.\/}, {Source code for ``Transferable Human Mobility Network Reconstruction with neuroGravity''} (2026). \url{https://doi.org/10.5281/zenodo.19727409}.

\end{thebibliography}
\end{document}